\documentclass[10pt,journal,compsoc]{IEEEtran}
\usepackage{times}
\usepackage{latexsym}
\usepackage{amsmath,amscd,amsbsy,amssymb,url,bm,amsthm}
\usepackage{enumitem,balance,mathtools}
\usepackage{wrapfig}
\usepackage{mathrsfs, euscript}
\usepackage{enumerate}
\usepackage{algorithmic}
\usepackage{multicol}
\usepackage{multirow}
\usepackage{booktabs}
\usepackage[vlined,ruled,commentsnumbered,linesnumbered]{algorithm2e}
\usepackage [colorlinks,linkcolor=blue] {hyperref}

\ifCLASSOPTIONcompsoc
  \usepackage[nocompress]{cite}
\else
  \usepackage{cite}
\fi

\ifCLASSINFOpdf
  \usepackage[pdftex]{graphicx}
\else
  \usepackage[dvips]{graphicx}
\fi

\usepackage{amsmath}
\usepackage{algorithmic}

\begin{document}
\title{Cross-lingual Universal Dependency Parsing \\ Only from One Monolingual Treebank}

\author{Kailai Sun$^\dag$, Zuchao Li$^\dag$, and Hai Zhao$^*$
	% <-this % stops a space
	\IEEEcompsocitemizethanks{\IEEEcompsocthanksitem{This paper was partially supported by National Key Research and Development Program of China (No. 2017YFB0304100), Key Projects of National Natural Science Foundation of China (U1836222 and 61733011), Huawei-SJTU long term AI project, Cutting-edge Machine Reading Comprehension and Language Model ($^\dag$Equal Contribution. $^*$Corresponding author: Hai Zhao).}
	\IEEEcompsocthanksitem{K. Sun, Z. Li, and H. Zhao are with the Department of Computer Science and Engineering, Shanghai Jiao Tong University, and also with Key Laboratory of Shanghai Education Commission for Intelligent Interaction and Cognitive Engineering, Shanghai Jiao Tong University, and also with MoE Key Lab of Artificial Intelligence, AI Institute, Shanghai Jiao Tong University. E-mail: \{kaishu2.0,charlee\}@sjtu.edu.cn, zhaohai@cs.sjtu.edu.cn. \protect}
	}
}

\markboth{Arxiv, April~2021}%IEEE Transactions on Pattern Analysis and Machine Intelligence
{Sun \MakeLowercase{\textit{et al.}}: Cross-lingual Universal Dependency Parsing Only from One Monolingual Treebank}

\IEEEtitleabstractindextext{%
\begin{abstract}
Syntactic parsing is a highly linguistic processing task whose parser requires training on treebanks from the expensive human annotation. As it is unlikely to obtain a treebank for every human language, in this work, we propose an effective cross-lingual UD parsing framework for transferring parser from only one source monolingual treebank to any other target languages without treebank available. To reach satisfactory parsing accuracy among quite different languages, we introduce two language modeling tasks into dependency parsing as multi-tasking. Assuming only unlabeled data from target languages plus the source treebank can be exploited together, we adopt a self-training strategy for further performance improvement in terms of our multi-task framework. Our proposed cross-lingual parsers are implemented for English, Chinese, and 22 UD treebanks. The empirical study shows that our cross-lingual parsers yield promising results for all target languages, for the first time, approaching the parser performance which is trained in its own target treebank.
\end{abstract}

}

\maketitle

\IEEEdisplaynontitleabstractindextext

% \fi
\IEEEpeerreviewmaketitle

\IEEEraisesectionheading{\section{Introduction}\label{sec:introduction}}


\IEEEPARstart{D}{ependency} syntactic parsing is a fundamental natural language processing task that discloses syntactic relations between words in a sentence. A neural parser usually consists of two parts: an encoder that transforms input text sequences into contextual representations and a decoder that generates the needed parse tree. Graph-based models \cite{mcdonald2005online,koo-collins-2010-efficient,dozat-etal-2017-stanfords} and transition-based models \cite{nivre2008algorithms} are the most typical solutions to this end. 

The syntactic annotations are designed and proposed by humans; making high-quality labeled data relies on human expertise, which is very time-consuming and painful. Therefore, although there is relatively sufficient labeled data (i.e., treebank) in some languages such as English, and dependency parsers demonstrate good performance in these languages \cite{ji-etal-2019-graph,zhang-etal-2020-efficient}, there are still many languages that lack manually annotated data.

To meet the huge demand of training syntactic parser among various languages, the project of universal dependencies (UD) treebanks was launched \cite{mcdonald-etal-2013-universal}. Though great efforts have been made as dozens of multilingual treebanks were annotated in terms of the project, it is still far from the huge demand of parsing thousands of human languages if we account for annotating treebanks to build parsers. Thanks to the remarkable progress in deep learning on language modeling, researchers can build powerful contextualized language models without any labeled text data. This motivates us to incorporate language modeling and cross-lingual transferring method for universal dependency parsing, which differs from the existing UD parsing or many cross-lingual parsing that we completely release the prerequisite of treebank for any target languages.

Cross-lingual parsing has been explored in previous studies, which focus on sharing limited word-level information by using multilingual word embeddings \cite{guo2015cross,tackstrom2012cross,ahmad-etal-2019-difficulties}, the parsing performance is quite unsatisfactory and still far below what is expected as long as treebank in the target language is unavailable, let alone the transferring parser being a practical tool for the use in the target language. To reach such an ambitious goal, letting the performance of cross-lingual universal dependency parsing approach that of parsers training on target language treebank, we borrow the ideas from successful practice in the pre-trained language models which learns broad and general knowledge from unlabeled language data and self-training which is capable of fully exploiting unlabeled structural data. By closely incorporating such two modeling methods, we propose a highly-effective cross-lingual universal parsing framework.

Pre-trained language models show that the generic nature of human languages can be effectively captured through language model training, which is also essentially important and helpful in cross-lingual parsing. Thus we introduce two language modeling training objectives, Word Ordering Modeling \cite{nishida2017word,wang2018improved} and Masked Language Modeling \cite{devlin-etal-2019-bert} as the auxiliary task for the dependency parsing, so that a multi-task learning framework is adopted for both training objectives of the language model and the original dependency parser.
As we do not assume there is a target language treebank available anymore, our training data will only consist of one source language treebank plus unlabeled data from target languages. We adopt a self-training method for better exploiting unlabeled data in our multi-task UD parser framework.

Our proposed cross-lingual parsers are implemented for English, Chinese, and 22 UD treebanks. The empirical study shows that our parsers transferred from only one monolingual treebank yield promising results for all target languages, for the first time, approaching the parser performance which is trained in its own target treebank. Besides, our proposed multi-task framework also generally helps monolingual supervised-learning parsers to give better performance at the same time.

\section{Related Work}

\noindent\textbf{Unsupervised Cross-lingual Parsing} Previous studies usually have to seek better parsing performance by adding cross-lingual features at word representation. Typical methods include removing all lexical features from the source treebank \cite{zeman2008cross,mcdonald2013universal}, selecting the underlying feature model from the Universal POS Tagset \cite{petrov-etal-2012-universal}, using cross-lingual word clusters \cite{tackstrom2012cross}, lexicon mapping \cite{xiao2014distributed,guo2015cross} and using relative position representation in order to ignore some language-unique features such as word order \cite{ahmad-etal-2019-difficulties}. However, unlike the above works, our focus is not on the word representation level. Instead, we go deep into the encoder by introducing additional language model training objectives so that the resulted model is capable of learning multiple objectives including syntactic parse in a contextualized way.

More closely related to our method, another pool of prior work focus on adapting the training process to better fit the target languages, including choosing the source language data points suitable for the target language \cite{sogaard2011data,tackstrom-etal-2013-target}, transferring from multiple sources \cite{mcdonald-etal-2011-multi,guo2016representation,tackstrom-etal-2013-target} and allowing unlabeled data from one or more auxiliary languages other than the source language to train the model \cite{ahmad-etal-2019-cross}. This method also used unannotated data. However, it only added a classifying languages objective which ignores a lot of inner-sentence knowledge of the unannotated data that is worth extracting. Our language model training objectives can effectively capture the generic nature of human languages from unlabeled data, and with self-training, the unlabeled data can be further exploited.

\noindent\textbf{Multilingual Representation Learning} The basic of the unsupervised cross-lingual parsing is that we can align the representations of different languages into the same space, at least at the word level. The main idea is that we can train a model on top of the source language embeddings, which are aligned to the same space as the target language embeddings, and thus all the model parameters can be directly shared across languages. In our work, we use Fasttext as our multilingual word embedding. This idea is further extended to learn multilingual contextualized word representations, for example, multilingual BERT \cite{devlin-etal-2019-bert}. Li et al. (2020) \cite{li2020global} shows that further improvements can be achieved when the encoders are trained on top of multilingual BERT. Thus we also apply multilingual BERT to our parser and because our language model training objectives in multi-tasking are more inclined to the target language, our improvements are more dramatical.

As our work is related to both syntactic parsing and language modeling, it is also related to those previous multi-task models which involve both of these two factors. Zhou et al. (2020b) \cite{2020LIMIT} presented LIMIT-BERT for learning language representations across multiple linguistics tasks, including dependency parsing with multi-task learning. However, our purpose is not the improvement of pre-training models. Instead, we focus on the cross-lingual dependency parsing performance by introducing several language model objectives into its training process. Our method is also not in conflict with pre-training models like LIMIT-BERT, and our proposed cross-lingual parser can be furthermore enhanced by incorporating the large-scale pre-training as it does.

\section{Biaffine Dependency Parser}\label{sec:dep_par}

Our method is based on the state-of-the-art graph-based deep biaffine dependency parser \cite{DBLP:journals/corr/DozatM16}.
For encoder, the biaffine parser applies multi-layer bidirectional LSTMs (BiLSTM) to encode the input sentence $[x_1,x_2,...,x_{l}]$. Each word-level representation $x_i$  is the concatenation of word/char/POS embeddings, i.e., $x_i=e_{w_i}\oplus e_{c_i}\oplus e_{p_i}$. The character-level
embedding $[e_{c_1},e_{c_2},...,e_{c_l}]$ is learned by convolutional neural networks (CNNs) to handle out-of-vocabulary words. The encoder outputs a sequence of context-dependent representations $[h_1,h_2,...,h_l]$.
$$h_i = \mathbf{BiLSTM}([e_{w_i};e_{c_i};e_{p_i}])$$
In order to distinguish the representation of words between head and dependent, they feed $h_i$ into two separate MLPs to get two lower-dimensional representation of the word, as a head and a dependent respectively.
$$r_i^{m} = \mathbf{ReLU}(\mathbf{MLP}^{m}(h_i)),m\in [head,dep]$$
The scores of all possible head-dependent pairs are computed via the Variable-class biaffine classifier \cite{DBLP:journals/corr/DozatM16}:
\begin{align*}
& R_{m}=[r_1^{m};r_2^{m};...;r_l^{m}],m\in [head,dep] \\
& S^{arc} = R^{T}_{head}U_{1}R_{dep}+u^{T}_{2}R_{head}+u^{T}_{3}R_{dep}+b.
\end{align*}
Similarly, the parser uses the other two MLPs to get the representation of the head and dependent for computing label scores through the fixed-class biaffine classifier.

In the training phase, the parser aims to optimize the following probability:
$$P_{\theta}(Y|X)=\prod \limits_{i=1}^l P_{\theta}(y_i^{label}|x_i,y_i^{arc})P_{\theta}(y_i^{arc}|x_i),$$
where $Y$ is a dependency tree, $X$ is the given sentence, $\theta$ denotes the learnable parameters and $y_i^{arc},y_i^{label}$ denote the gold-standard head and dependency relation for word $x_i$. The training objective for parser is the cross-entropy loss, which minimizes the negative log-likelihood:
\begin{align*}
\begin{aligned}
\mathcal{L}^{parse}= -\sum \limits_{i=1}^l \big(\log P_{\theta}(y_i^{arc}|x_i) + \log P_{\theta}(y_i^{label}|x_i,y_i^{arc})\big).
\end{aligned}
% \mathcal{L}^{arc}+\mathcal{L}^{label} \\
%\mathcal{L}^{arc}=
% &\mathcal{L}^{label}=
\end{align*}

Denote $DL$ as the set of dependency label. In the inference phase, the graph-based dependency parser uses a max spanning tree (MST) decoder to find the highest-scoring tree relying on all $p_{i,j}^{arc}$: the probability that $x_j$ is the head of $x_i$ and $p_{i,j,k}^{label}$: the probability that $d_k$ is the label of $(x_i,x_j)$ if $(x_i,x_j)$ is an arc. ($i,j\in \{1,2,...,l\},d_k \in DL$).

\begin{figure*}[]
  \centering
  \includegraphics[scale=0.5]{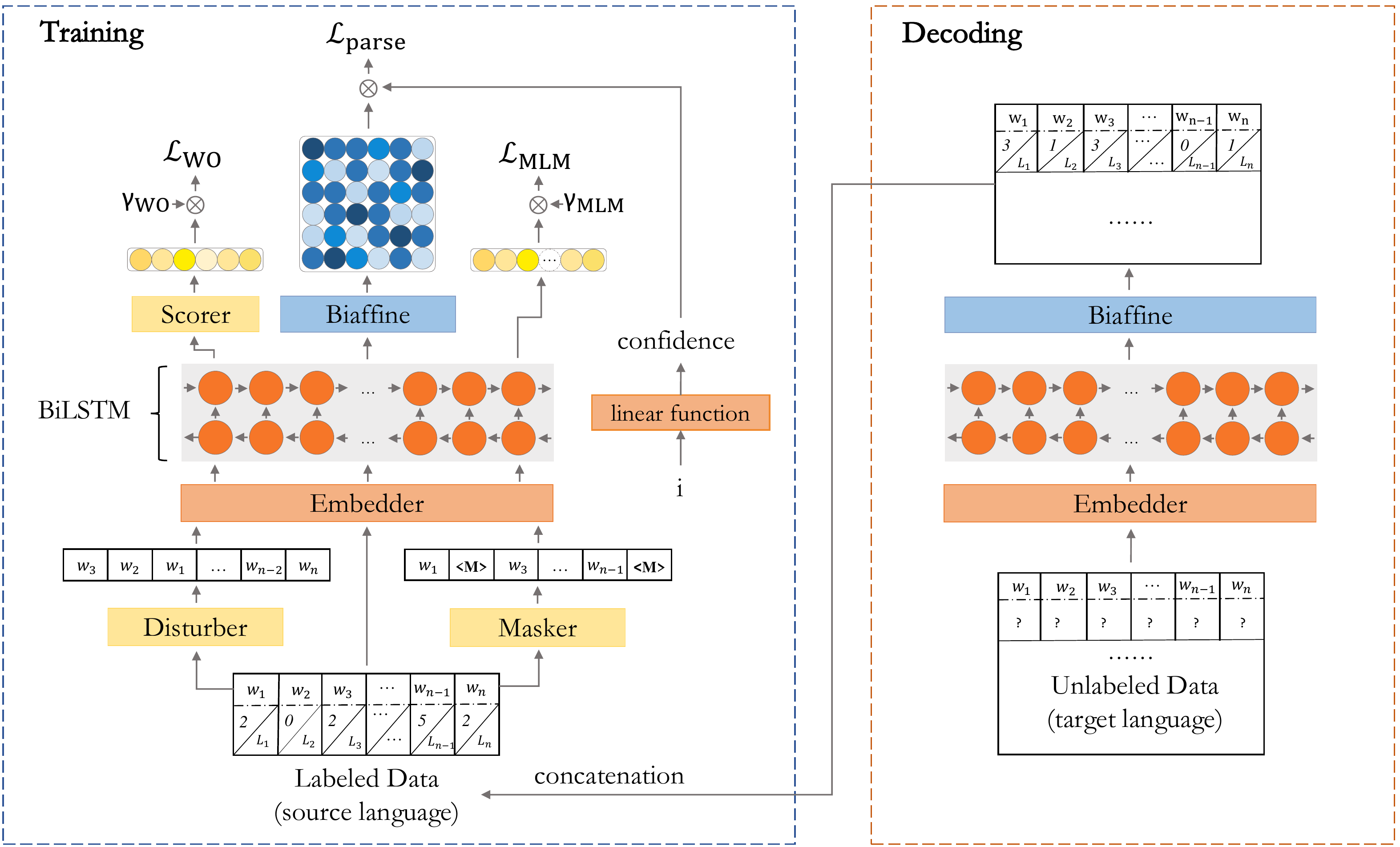}
  \caption{Multi-tasking and self-training.}\label{fig:network}
\end{figure*}

\section{Methodology}

In this section, we introduce two language modeling pre-training tasks we adopted and present how to train the multi-tasking framework in a self-training way.

\subsection{Language Model Training Objective}
To capture general language representation only from plain text, we need to build a language model (LM) through discriminative training objectives. Thus we adopt two LM training objectives as our enhanced learning objective to help syntactic learning, masked language model (MLM), and word ordering (WO). Especially the latter directly models the word order which expresses grammatical relationships in many languages. Because different languages have different word order preferences, word order is an important factor that makes cross-lingual syntactic parsing difficult, and it is a barrier that cannot be ignored in cross-lingual tasks.

\noindent\textbf{Masked Language Modeling.} For this training objective, we randomly choose 15\% of the words in the sentence. Among the selected words, 80\% of them are replaced by [MASK], 10\% keep their original tokens, and the remaining 10\% are represented by random tokens. Then we feed the pretreated sentence into multiple layers BiLSTM encoder and map the output of the encoder into word alphabet dimension for predicting those masked words. MLM models the relationships between words and words that are based on the syntax, so it's potentially helpful for syntactic parsing.

\noindent\textbf{Word Ordering} has become a common pre-training task, and today it has been structured in a variety of ways. In WO, the order of some words in the sentence is randomly changed and the model is asked to restore the sentence to its original order. In this paper, we built our WO model referring to Nishida and Nakayama (2017) \cite{nishida2017word}. 

We use multi-layer BiLSTM with a soft attention mechanism to encode the input sentence $[x_1,x_2,...,x_{l}]$. The BiLSTM's $t$-th hidden state $h_t$ and memory cells $c_t$ are computed by the following function:
\begin{equation*}
    \begin{aligned}
        \overline{x}&=\sum_{i=0}^{l}x_i/l \\
    	h_t,c_t &= \begin{cases}
    	\mathbf{MLP}(\overline{x}),0 &\;(t=0)\\
    	F_{\mathbf{LSTM}}(x_{i_{t-1}},h_{t-1},c_{t-1}) &\;(1\leq t\leq l),
    	\end{cases}
    \end{aligned}
\end{equation*}
where $h_0$ is obtained by feeding the average of all the word representations into an MLP. The function $F_{\mathbf{LSTM}}$ is the state-update function of BiLSTM and $i_{t-1}\in \{1,...,l\}$ represents the index of word in the input sentence which is predicted to be placed at position $t-1$. Denote $p_{t,i}$ as the probability of word $w_i$ to be placed at the $t$-th position. $i_{t-1}$ can be computed as follows:
\begin{align*}
& i_{t-1} = \mathop{\mathbf{argmax}}\limits_{i\in \{1,...,l\}}\ p_{t-1,i} \\
& p_{t,i}=\frac{exp(score(h_t,x_i))}{\sum_{k=0}^{l}exp(score(h_t,x_k))}.
\end{align*}
We use a variable-class biaffine classifier as the function $score$ to compute the confidence of placing word $w_i$ at position $t$.
$$score(h_t,x_i)=U_s(\mathbf{Tanh}(h_t+x_i))$$
%where $U_s\in R^{hid-dim\times 1}$.
We use cross-entropy loss which is as same as the loss function of dependency parser as the training objective for WO:
\begin{align*}
&\mathcal{L}^{wo}=-\sum \limits_{t=1}^l \log p_{t,i_{t}} \\
& p_{t,i_{t}}=P_{\theta}(x_{i_{t}}|x_{i_1},x_{i_2},...,x_{i_{t-1}}).
\end{align*}

\subsection{Multi-task Learning}

We train the above two language modeling objectives and dependency parsing with a multi-task diagram. These three models share the same multi-layer BiLSTM encoder and use their individual decoder for getting their own objectives. In this way, the overfitting of the encoder on the source training set can be avoided to a certain extent, and the context information that is more general and helpful to the target set can be learned. To further enhance the word-level representations, we joint a word representation $e_{lm}$ from pre-trained XLNet \cite{DBLP:conf/nips/YangDYCSL19} or BERT \cite{devlin-etal-2019-bert} model. 
\begin{align*}
h_i^{m} = \mathbf{BiLSTM}([e_{w_i}^{m};e_{c_i}^{m};e_{p_i}^{m};e_{lm_i}^{m}]) \\ m \in \{parse,wo,mlm\}    
\end{align*}

It is obvious that modeling the unlabeled data of the target language is helpful for cross-lingual syntactic parsing. However, only using the source language data for training parsing is not the best choice. We find out from the experiment that it is also helpful to model the source language data at the same time.

Since the training set for parsing is much smaller than which for traditional pre-training model, it is very difficult for our WO and MLM to fully converge when accompanying with parsing objective. So we add two weights $\gamma_{wo}$ and $\gamma_{mlm}$ to the loss of these two objectives. The training loss of the multi-task is calculated by weighting the losses of the three objectives together.
$$\mathcal{L}=\mathcal{L}^{parse}+\gamma_{wo} \mathcal{L}^{wo}+\gamma_{mlm}\mathcal{L}^{mlm}$$
The overall structure of our multi-task learning is shown on the left side of Figure \ref{fig:network}. We only use the dependency parsing decoder in the inference phase, which is as same as the one we introduced in Section \ref{sec:dep_par}.

\begin{algorithm}[t]
  \KwIn{source training data $D_{source}=(X_{source},Y_{source})$,\\ \hspace*{2.8em} extra target language data\\ \hspace*{2.8em} $D_{target} =(X_{target})$}
  \KwOut{the ensemble model of round $n$}
  \BlankLine
  \caption{Self-training Process}\label{Alg_Quick}
  teacher-model $\leftarrow$ None\;
  \For{$i \leftarrow 1$ \KwTo $n$}{
    \If{not teacher-model}{
      set different random seed to train $m_i$ student models $M_{i}=\{m_{i1},...,m_{im_i}\}$ on $D_{source}$\;
    }
    \Else{
     $\textnormal{Conf}_i \leftarrow$ $\alpha_c i + \beta_c$\; 
     $\Tilde{Y}_{target}^{i} \leftarrow$ teacher-model$(X_{target})$\;
     $D_{train}=D_{source}\cup (X_{target},\Tilde{Y}_{target}^{i},\textnormal{Conf}_i)$\;
     set different random seed to train $m_i$ student models $M_{i}=\{m_{i1},...,m_{im_i}\}$ on $D_{train}$\;
    }
    teacher-model $\leftarrow$ ensemble models in $M_{i}$\;
  }
  $\Return$ teacher-model\;
\end{algorithm}

\begin{table*}[t!]
    \caption{The monolingual and cross-lingual parsing results on 22 UD treebanks. $^\dag$ denotes the results from BIAF models trained by ourselves.}\label{tab:ud} % 
    \centering
        \begin{tabular}{l|cc|cc|cc||cc|cc|cc}
        \toprule
         & \multicolumn{6}{c||}{\bf Monolingual} & \multicolumn{6}{c}{\bf Cross-lingual}\\
        \cmidrule(lr){2-7}\cmidrule(lr){8-13} & \multicolumn{2}{c|}{BIAF\cite{ma-etal-2018-stack}} & \multicolumn{2}{c|}{\bf Ours} & \multicolumn{2}{c||}{\bf Ours +BERT} &  \multicolumn{2}{c|}{Ahmad et al.\cite{ahmad-etal-2019-difficulties}} & \multicolumn{2}{c|}{\bf Ours} & \multicolumn{2}{c}{\bf Ours +BERT} \\ 
        & UAS & LAS & UAS & LAS & UAS & LAS & UAS & LAS & UAS & LAS & UAS & LAS \\
        \midrule
        $bg$ & 94.30 & 90.04 & \textbf{94.44} & \textbf{90.41} & \textbf{94.73} & \textbf{90.88} & 74.85 & 65.01 & \textbf{82.76} & \textbf{68.77} & \textbf{88.29} & \textbf{73.46}\\ 
        $ca$ & 94.36 & 92.05 & \textbf{94.89} & \textbf{92.96} & \textbf{95.36} & \textbf{93.67} & 70.96 & 62.85 & \textbf{79.18} & \textbf{68.21} & \textbf{83.73} & \textbf{72.65}\\ 
        $cs$ & 94.06 & 90.60 & \textbf{94.38} & \textbf{91.44} & \textbf{94.38} & \textbf{91.52} & 59.56 & 51.80 & \textbf{67.07} & \textbf{54.40} & \textbf{78.64} & \textbf{65.49}\\ 
        $nl$ & \textbf{93.44} & \textbf{91.04} & 93.12  & 90.12 & \textbf{95.11} & \textbf{92.81} & 66.45 & 59.54 & \textbf{72.85} & \textbf{64.09} & \textbf{86.37} & \textbf{79.86}\\
        $en$ & 91.91 & 89.82 & \textbf{92.16} & \textbf{90.25} & \textbf{92.74} & \textbf{90.60} & 89.46 & 87.54 & $-$ & $-$ & $-$ & $-$ \\ 
        $et$ & 89.39$^\dag$ & 86.95$^\dag$ & \textbf{89.97} & \textbf{87.30} & \textbf{90.95} & \textbf{88.54} & 63.08 & 45.45 & \textbf{69.25} & \textbf{53.18} & \textbf{81.53} & \textbf{64.30}\\ 
        $fi$ & 89.87$^\dag$ & 88.16$^\dag$ & \textbf{91.25} & \textbf{88.33} & \textbf{92.47} & \textbf{89.87} & 65.04 & 49.98 & \textbf{71.15} & \textbf{56.21} & \textbf{83.46} & \textbf{65.28}\\ 
        $fr$ & 92.62 & 89.51 & \textbf{93.83} & \textbf{90.76} & \textbf{94.10} & 90.65 & 76.11 & 71.79 & \textbf{83.15} & 73.23 & \textbf{87.00} & \textbf{75.33}\\
        $de$ & \textbf{90.26} & \textbf{86.11} & 89.48  & 84.49 & \textbf{90.65} & \textbf{86.64} & 67.60 & 58.86 & \textbf{76.20} & \textbf{63.39} & \textbf{86.43} & \textbf{77.57} \\
        $he$ & 91.52$^\dag$ & 89.71$^\dag$ & \textbf{91.69} & \textbf{89.90} & \textbf{93.04} & \textbf{91.46} & 53.04 & 46.16 & \textbf{63.53} & \textbf{53.77} & \textbf{72.98} & \textbf{64.12}\\ 
        $hi$ & 95.78$^\dag$ & 93.25$^\dag$ & \textbf{96.15} & \textbf{93.58} & 95.91 & 93.05 & 30.94 & 23.55 & \textbf{45.56} & \textbf{39.40} & \textbf{63.28} & \textbf{57.57}\\ 
        $id$ & 88.26$^\dag$ & 83.02$^\dag$ & \textbf{88.42} & \textbf{83.17} & \textbf{90.09} & \textbf{84.81} & 47.08 & 42.78 & \textbf{59.86} & \textbf{53.39} & \textbf{69.43} & \textbf{61.25}\\ 
        $it$ & 94.75 & 92.72 & \textbf{95.16} & \textbf{93.41} & \textbf{95.84} & \textbf{94.05} & 78.63 & 74.31 & \textbf{83.35} & 75.62 & \textbf{87.95} & \textbf{77.02}\\
        $ko$ & 89.07$^\dag$ & 86.94$^\dag$ & \textbf{89.64} & \textbf{87.27} & \textbf{90.41} & \textbf{88.63} & 33.08 & 16.96 & \textbf{41.46} & \textbf{27.51} & \textbf{63.79} & \textbf{56.82}\\
        $la$ & 84.43$^\dag$ & 80.49$^\dag$ & \textbf{84.79} & \textbf{80.73} & \textbf{87.83} & \textbf{83.29} & 45.96 & 33.91 & \textbf{58.69} & \textbf{51.24} & \textbf{70.15} & \textbf{61.33}\\ 
        $lv$ & 87.30$^\dag$ & 82.89$^\dag$ & \textbf{87.59} & \textbf{83.10} & \textbf{90.32} & \textbf{87.22} & 66.95 & 49.66 & \textbf{74.61} & \textbf{56.04} & \textbf{85.34} & \textbf{68.27}\\ 
        $no$ & 95.28 & 93.58 & \textbf{95.30} & \textbf{93.93} & 93.27 & 91.05 & 78.47 & 71.50 & \textbf{83.52} & \textbf{72.85} & \textbf{87.07} & \textbf{79.94}\\ 
        $pl$ & 96.91$^\dag$ & 94.45$^\dag$ & \textbf{97.22} & \textbf{94.83} & 95.32 & 93.46 & 71.73 & 60.83 & \textbf{80.40} & \textbf{64.37} & \textbf{86.78} & \textbf{70.16}\\ 
        $ro$ & 91.94 & 85.61 & \textbf{92.33} & \textbf{86.09} & \textbf{92.70} & \textbf{86.59} & 61.19 & 51.45 & \textbf{67.92} & \textbf{54.81}  & \textbf{80.86} & \textbf{64.25}\\
        $ru$ & 94.40 & 92.68 & \textbf{94.90} & \textbf{93.49} & 94.52 & 92.35 & 55.40 & 47.84 & \textbf{65.70} & \textbf{54.96} & \textbf{75.97} & \textbf{63.46}\\
        $sk$ & 91.71$^\dag$ & \textbf{86.92}$^\dag$ & \textbf{91.74} & 86.89 & \textbf{93.21} & \textbf{88.41} & 63.66 & 56.38 & \textbf{70.29} & \textbf{61.87} & \textbf{82.42} & \textbf{69.95}\\
        $es$ & 93.72 & 91.33 & \textbf{94.27} & \textbf{92.34} & \textbf{94.81} & \textbf{93.01} & 71.50 & 64.40 & \textbf{78.84} & \textbf{68.98} & \textbf{85.11} & \textbf{73.54} \\
        \midrule
        \bf AVG & 92.06 & 88.99 & \textbf{92.40} & \textbf{89.31} & \textbf{93.08} & \textbf{90.11} & 63.22 & 54.21 & \textbf{70.25} & \textbf{58.87} & \textbf{80.32} & \textbf{68.65}\\ 
        \bottomrule
        \end{tabular}
\end{table*}

\subsection{Self-training}
Ensemble is the way in which we take the weighted average of the distributions predicted by several models to get further improvement. Suppose we have $n$ models $\theta_{1:n}$ and the $m$-th model predicts the distribution $p_m^{arc}$ and $p_m^{label}$, then the ensemble of $n$ models predicts the distribution $e^{arc}$ and $e^{label}$ as follows:
\begin{align*}
& e_{i,j}^{arc}=\sum \limits_{m=1}^{n}\alpha_m p_m^{arc}(x_j|x_i,\theta) \\
& e_{i,j,k}^{label}=\sum \limits_{m=1}^{n}\alpha_m p_m^{label}(d_k|x_i,x_j,\theta).
\end{align*}

In self-training, we made an ensemble of the models obtained from a single training round with different random seeds and use this ensemble model as the teacher model of the next training round. Instead of minimizing the cross-entropy based on the golden distribution, we use the teacher model's predict distribution $e_{i,j}^{arc}$ and $e_{i,j,k}^{label}$ as target and minimizes the loss:
\begin{align*}
& \mathcal{L}^{parse}_{SF}=-\sum \limits_{i=1}^{l}\sum \limits_{j=1}^{l} e^{arc}(x_j|x_i)\times \log P_{\theta}(x_j|x_i)\\ & -\sum \limits_{i=1}^{l}\sum \limits_{j=1}^{l}\sum \limits_{k=1}^{|DL|} e^{label}(d_k|x_i,x_j)\times \log P_{\theta}(d_k|x_i,x_j),
\end{align*}
where $\mathcal{L}^{parse}_{SF}$ denotes the self-training loss, $DL$ denotes the label set, $X$ denotes the given sentence, and $Y$ denotes a dependency tree. Rewrite the function to the sequence-level:
$$\mathcal{L}_{SF}=-\sum \limits_Y e(Y|X)\times \log P_{\theta}(Y|X).$$
Due to the exponential large search space of $Y$, we make an approximation of the loss function by replacing the teacher distribution $e$ with a one-hot distribution, which has the probability 1 on the parsing result $\Tilde{Y}$ of the teacher model. 
$$\mathcal{L}^{parse}_{SF}  \approx -\log P_{\theta}(\Tilde{Y}|X)$$

Algorithm \ref{Alg_Quick} shows the process of self-training where the source training data is noted as $D_{source}$, the extra unannotated data of target language is noted as $D_{target}$ and $D_{train}$ represents the training set. We iteratively train $n$ rounds of models and in the $i$th round, we train $m_i$ models. $[\{m_{1,1},...,m_{1,m_1}\};...;\{m_{n,1},...,m_{n,m_n}\}]$. We take the ensemble of $m_i$ models in round $i$ as the teacher model of round $i+1$. We use the teacher model to parse the extra unlabeled data of the target language and concatenate these data with the labeled source language data as the training set for the current training round. Since the data parsed by the teacher model is not completely correct, we give a confidence hyperparameter $\textnormal{Conf}$ for the extra data in each round. And since the accuracy of the teacher model increases with iteration, we take the value of the confidence hyperparameter $\textnormal{Conf}_{i}$ as an increacing linear function of the number of iterations $i$. 
The final loss function can be applied as follows:
$$\mathcal{L}=\mathcal{L}^{parse}(X_{source})+ \textnormal{Conf} \ \mathcal{L}^{parse}_{SF}(X_{target})$$
$$+ \gamma_{wo} \mathcal{L}^{wo}(X_{train})+\gamma_{mlm}\mathcal{L}^{mlm}(X_{train}).$$

In the process of iteration, the value of $\textnormal{Conf}$ keeps increasing, which makes the influence of teacher model on training becomes more and more important, leading to smaller and smaller differences in models obtained by using different seeds in the same round. Therefore, as the number of iterations increase, we keep reducing the number of models trained in one round, making full use of ensemble while avoiding unnecessary training process.

The models can be ensembled as long as they have the same parsing decoding strategy (MST). So we do not need each model to have the same structure. That is to say; we can use other language modeling objectives to replace WO or MLM in which makes our work worth further study. For example, we can increase the difference between models by using different language modeling objectives in order to increase the room for improvement of ensemble and then the life cycle of self-training can be extended.

\begin{table}[]
    \caption{Details of the selected languages. “IE” is the abbreviation of Indo-European.}\label{tab:lang_family}
    \centering
        \begin{tabular}{p{60pt}|p{50pt}|p{50pt}|r}
        \toprule
        Language & Lang.Family & Treebank & Sents\\
        \midrule
        Bulgarian (bg) & IE.Slavic & BTB & 8,907\\ 
        %\hline
        Catalan (ca) & IE.Romance & AnCora & 13,123\\ 
        %\hline
        Czech (cs) & IE.Slavic & PDT & 102,993 \\ 
        %\hline
        Dutch (nl) & IE.Germanic & Alpino & 18,058\\ 
        %\hline
        English (en) & IE.Germanic & EWT  & 12,543\\ 
        %\hline
        Estonian (et) & Uralic & EDT  & 20,827\\ 
        %\hline
        Finnish (fi) & Uralic & TDT & 12,217\\ 
        %\hline
        French (fr) & IE.Romance & GSD & 14,554 \\ 
        %\hline
        German (de) & IE.Germanic & GSD & 13,814 \\ 
        %\hline
        Hebrew (he) & Afro-Asiatic & HTB & 5,241 \\ 
        %\hline
        Hindi (hi) & IE.Indic & HDTB & 13,304\\
        %\hline
        Indonesian (id) & Austronesian & GSD & 4,477\\
        %\hline
        Italian (it) & IE.Romance & ISDT & 13,121\\
        %\hline
        Korean (ko) & Korean & GSD & 27,410\\
        %\hline
        Latin (la) & IE.Latin & PROIEL & 15,906\\
        %\hline
        Latvian (lv) & IE.Baltic & LVTB & 5,424\\
        %\hline
        Norwegian (no) & IE.Germanic & Bokmaal & 29870\\
        %\hline
        Polish (pl) & IE.Slavic & LFG & 19,874\\
        %\hline
        Romanian (ro) & IE.Romance & RRT & 8,043\\
        %\hline
        Russian (ru) & IE.Slavic & SynTagRus & 48,814\\
        %\hline
        Slovak (sk) & IE.Slavic & SNK & 8,483\\
        %\hline
        Spanish (es) & IE.Romance & AnCora & 28,492\\
        \bottomrule
        \end{tabular}
\end{table}

\section{Experiments}
We evaluate our method using the Universal Dependencies
(UD) Treebanks (v2.2) \cite{d701bee1cabe492caf36340d6341e27b}. We select 22 languages which are shown in Table \ref{tab:lang_family}.
We take English as the source language and the other 21 languages as the target languages. The unlabeled data for 21 target languages come from Wikipedia, with the selection criterion being that the sentence contains more than 10 words. We use UDPipe \cite{straka-strakova-2017-tokenizing} to generate the Universal POS tag for the unlabeled data. We evaluate the monolingual performance of our model on the English Penn Treebank (PTB) \cite{marcus-etal-1993-building}, the Chinese Penn Treebank (CTB) \cite{xue-etal-2002-building} and the same 22 languages from UD.

We use GloVe \cite{pennington2014glove} and BERT-\textit{large-uncased} model for PTB, and FastText \cite{bojanowski2017enriching} and BERT-\textit{base-chinese} model for CTB. For UD, we use FastText and BERT-\textit{base-multilingual-cased} model. CLS representations is used as the ROOT embedding when Bert is used and we randomly initialize the ROOT embedding when Bert is not used.

Unlabeled Attachment Scores (UAS) and Labeled Attachment Scores (LAS) are adopted as the evaluation metrics. For PTB and CTB, punctuation is excluded as in previous work \cite{DBLP:journals/corr/DozatM16}. While for UD, punctuation is included as well as the standard evaluation criteria.

\subsection{Implementation Details}

\noindent\textbf{Hyperparameters.} In character CNN, the convolutions have a window size of 3 and consist of 50 filters. We use 3 bidirectional LSTMs with 512-dimensional hidden states as the encoder. 
In our parser, the outputs of the BiLSTM employ a 512-dimensional MLP layer for the arc scorer and a 128-dimensional MLP layer for the label scorer with all using ReLU as the activation function. In WO, the learnable linear function map the embedding into 512-dimensions. %In Pointer Network-based WO, we use 1 layer BiLSTM with a 512-dimensional hidden size as the decoder.

\noindent\textbf{Training.} Parameter optimization is performed with the Adam optimizer with $\beta_1 = \beta_2 = 0.9$. We choose an initial learning rate of $\eta_0 = 0.001$. The learning rate $\eta$ is annealed by multiplying a fixed decay rate $\rho = 0.999995$ when parsing performance stops increasing on validation sets. 
To reduce the effects of an exploding gradient, we use a gradient clipping of 5.0. 
For the BiLSTM in encoder and decoder, we use recurrent dropout with a drop rate of 0.33 between hidden states and 0.33 between layers. 
Following Dozat and Man-ning (2017) \cite{DBLP:journals/corr/DozatM16}, we also use embedding dropout with a rate of 0.33 on all word, character, and POS tag embeddings. 
The weights for the losses of WO and MLM are 0.2 and 0.15.

\noindent\textbf{Self-training.} 
The number of models in each training round are 5,5,4,3,2,2,...2. The parameters of the increasing linear function for computing $\textnormal{Conf}$ are $\alpha_c=0.6,\beta_c=0.03$ when the target language and the source language are in the same language family and $\alpha_c=0.4,\beta_c=0.05$ when they are in different language families. We terminate the iteration when the increase between two rounds is less than $0.2\%$ and the number of iterations should not exceed 8.

\begin{figure}[]
  \centering
  \includegraphics[scale=0.48]{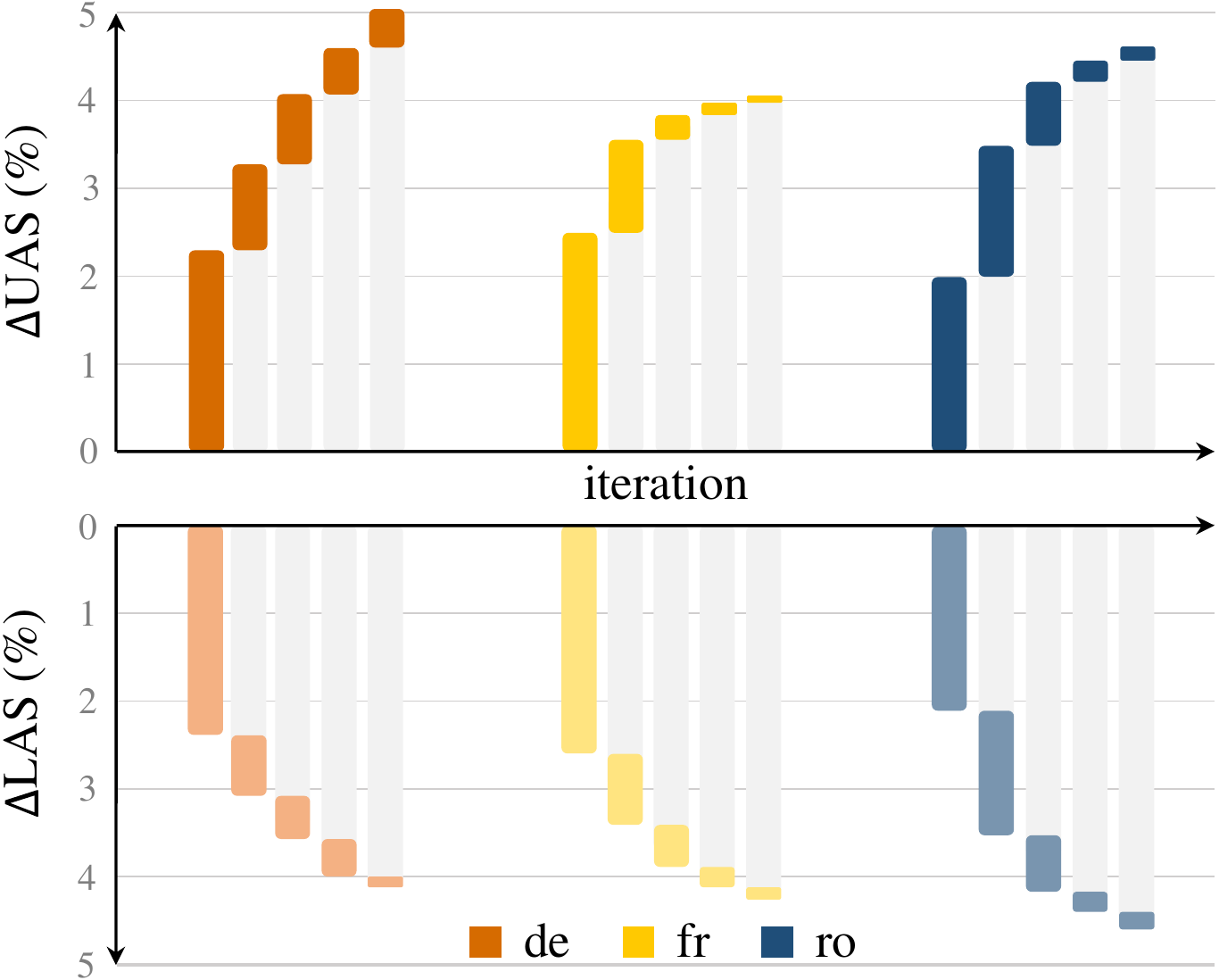}
  \caption{The increase of UAS and LAS vs. iterations. The top half plots the increase of UAS vs. iterations while the bottom half plots the increase of UAS vs. iterations.}\label{fig:growth_uas_las}
\end{figure}

\subsection{Main Results}

For zero-shot cross-lingual transfer learning, we take English as the source language and other 21 UD languages as the target languages. 
We applied self-training to our multi-tasking model. In the first round, we only use the training set of $en$ to train the model, which means no additional unlabeled data has been used. 
Starting from the second round, we ensemble a range of models we gained from the previous round to annotate the unlabeled data, which will be concatenated to the training set in the next round. The amount of unlabeled data that we use for each target language is 15K. For each language, we performed self-training process five times with different random seeds, using the $en$ development set to select the optimal model, and reporting the average results on the target language test set.

The right hand of Table \ref{tab:ud} shows the average results of our method. For comparison, we also show the RNN-Graph-based results reported by Ahmad et al. (2019a) \cite{ahmad-etal-2019-difficulties}. It can be seen that our method results in significant performance improvements in cross-lingual transfer learning from only one monolingual treebank ($en$) to all 21 target languages. The results of the Indo-european languages especially which in IE.Germanic are outstanding being attribute to the high similarity between these languages and English. The results of Ahmad et al. (2019a) \cite{ahmad-etal-2019-difficulties} also imply this conclusion. Although the baselines for languages in other families are lower,  fortunately for these languages our method offers greater room for improvement which have an average of 8.79\% compared with an average of 6.48\% for Indo-european languages.

With the implementation of multilingual BERT, our method yields promising results for each target language, for the first time, approaching the parser performance trained in its own target treebank.

To better illustrate our self-training process, we select three languages, $de$, $fr$ and $ro$, and plot the growth value of our parser's UAS and LAS, denoted by $\triangle$UAS and $\triangle$LAS, on their test sets during self-training in Figure \ref{fig:growth_uas_las}. As the number of iterations increases, the growth rate gradually decreases until the dynamics stabilize. This seems to indicate that this self-training also has a performance bottleneck, but compared to unsupervised, this bottleneck is obviously higher.

\begin{table}[]
    \caption{Cross annotation standard results from $fr$, $de$ and $it$ UD treebanks to PTB.}\label{tab:cross_stand}
    \centering
        \begin{tabular}{l|cc|cc|cc}
        \toprule
        \multirow{2}{*}{Methods} & \multicolumn{2}{c|}{$fr$} & \multicolumn{2}{c|}{$de$} & \multicolumn{2}{c}{$it$} \\ 
        & UAS & LAS & UAS & LAS & UAS & LAS\\
        \midrule
        Right-arc Baseline  & 18.23 & $-$ &  18.23 & $-$ & 18.23 & $-$ \\ 
        Ours  & 53.46 & 32.15 & 50.93 & 30.47 & 46.14 & 29.86 \\ 
        \bottomrule
        \end{tabular}
\end{table}

Additionally, our method is not limited to cross-lingual transfer learning under the same annotation standard. For cross annotation standard case, our method can still train a parser with competitive accuracy on the target language. We take 3 UD treebanks, $fr$, $de$, $it$, respectively, as the source language and English as the target language whose annotation standard is based on English Penn Treebank. We use the training set of PTB as the unlabeled data. Table \ref{tab:cross_stand} shows our final score on PTB test set under the cross-annotation-standard case. We also show the right-arc baseline of PTB test set in which we simply set the head of a word as the word right before it and the head of the first word is ROOT. Compared with the right arc baseline, our method has achieved consistent improvement in the three source languages, which shows that our 
self-training method can effectively capture syntactic information. In addition, due to the similarity of the syntax annotations of each source language and the difference between source and target languages, the final improvement effect of our method also differs.

\subsection{Monolingual Results}

Although our model is designed to be a supposed cross-lingual parser, it can be applied to the monolingual case. For monolingual case, we only train the model in one round with no additional unlabeled data. 
\begin{table}[]
    \caption{Monolingual performance on PTB and CTB benchmarks. BERT is denoted by B, XLNet is denoted by X.}\label{tab:ptb_ctb}
    \centering
        \begin{tabular}{l|cc|cc}
            \toprule
            \multirow{2}{*}{Systems} & \multicolumn{2}{c|}{PTB} & \multicolumn{2}{c}{CTB} \\ 
            & UAS & LAS & UAS & LAS \\
            \midrule
            Dozat and Manning \cite{DBLP:journals/corr/DozatM16} & 95.74  & 94.08 & 89.30 & 88.23 \\ 
            %\hline
            Ma et al. (2018) \cite{ma-etal-2018-stack} & 95.87  & 94.19 & 90.59 & 89.29 \\ 
            %\hline
            Ji et al. (2019b) \cite{ji2019graph} & 95.87  & 94.15 & 90.78 & 89.50 \\ 
            %\hline
            Zhou and Zhao (2019) \cite{zhou-zhao-2019-head} & \textbf{96.09} & \textbf{94.68} & 91.21 & 89.12 \\ 
            %\hline
            \bf Ours  & 96.05 & 94.43  & \textbf{91.22} & \textbf{90.05}  \\ 
            \midrule
            \multicolumn{5}{c}{w/ Pre-training}  \\ 
            \midrule
            Li et al. (2020) \cite{li2020global} +B $^*$ & 96.55 & 94.70 & $-$  & $-$  \\ 
            Wang and Tu (2020) \cite{wang-tu-2020-second} +B & 96.91 & 95.34 & 92.55  & 91.38  \\ 
            %\hline
            Zhou et al. (2020a) \cite{zhou-etal-2020-parsing} +B & 96.90 & 95.32 & $-$ & $-$  \\ 
            %\hline
            Zhou and Zhao (2019) \cite{zhou-zhao-2019-head} +B & 97.00 & 95.43 & $-$ & $-$  \\ 
            %\hline
            Zhou et al. (2020b) \cite{2020LIMIT} +B & \textbf{97.14} & \textbf{95.44}  & $-$ & $-$  \\ 
            %\hline
            \bf Ours +B  & 96.93  & 95.28  & \textbf{92.73} & \textbf{91.50}  \\ 
            \bf Ours +B $^*$  & 96.66  & 95.10  & \textbf{92.65} & \textbf{91.38}  \\
            %\hline
            Zhou et al. (2020a) \cite{zhou-etal-2020-parsing} +X & \textbf{97.23} & \textbf{95.65} & $-$ & $-$  \\ 
            %\hline
            \bf Ours +X  & 97.07  & 95.48  & $-$ & $-$  \\ 
            \bottomrule
        \end{tabular}
\end{table}

Table \ref{tab:ptb_ctb} shows the monolingual results of our method as well as the reported scores of current state-of-the-art works on PTB and CTB. Our model achieves state-of-the-art performance on CTB and gains almost the same performance as Zhou and Zhao (2019) \cite{zhou-zhao-2019-head} on PTB. Simultaneously, in the case of applying the pre-trained language model, our model's performance is comparable to the current state-of-the-art works. Since the baseline score on PTB is very higher, to show the true improvement of our proposed method, we performed a significance test on PTB by randomly sampling 50 sentences on the test set 1000 times, and evaluating the UAS and LAS scores on these sub-test sets respectively. The final conclusion is that our results are higher than the baseline at the significance level $p<0.05$, indicating that our improvement is significant.

\begin{figure}[h]
  \centering
  \includegraphics[scale=0.35]{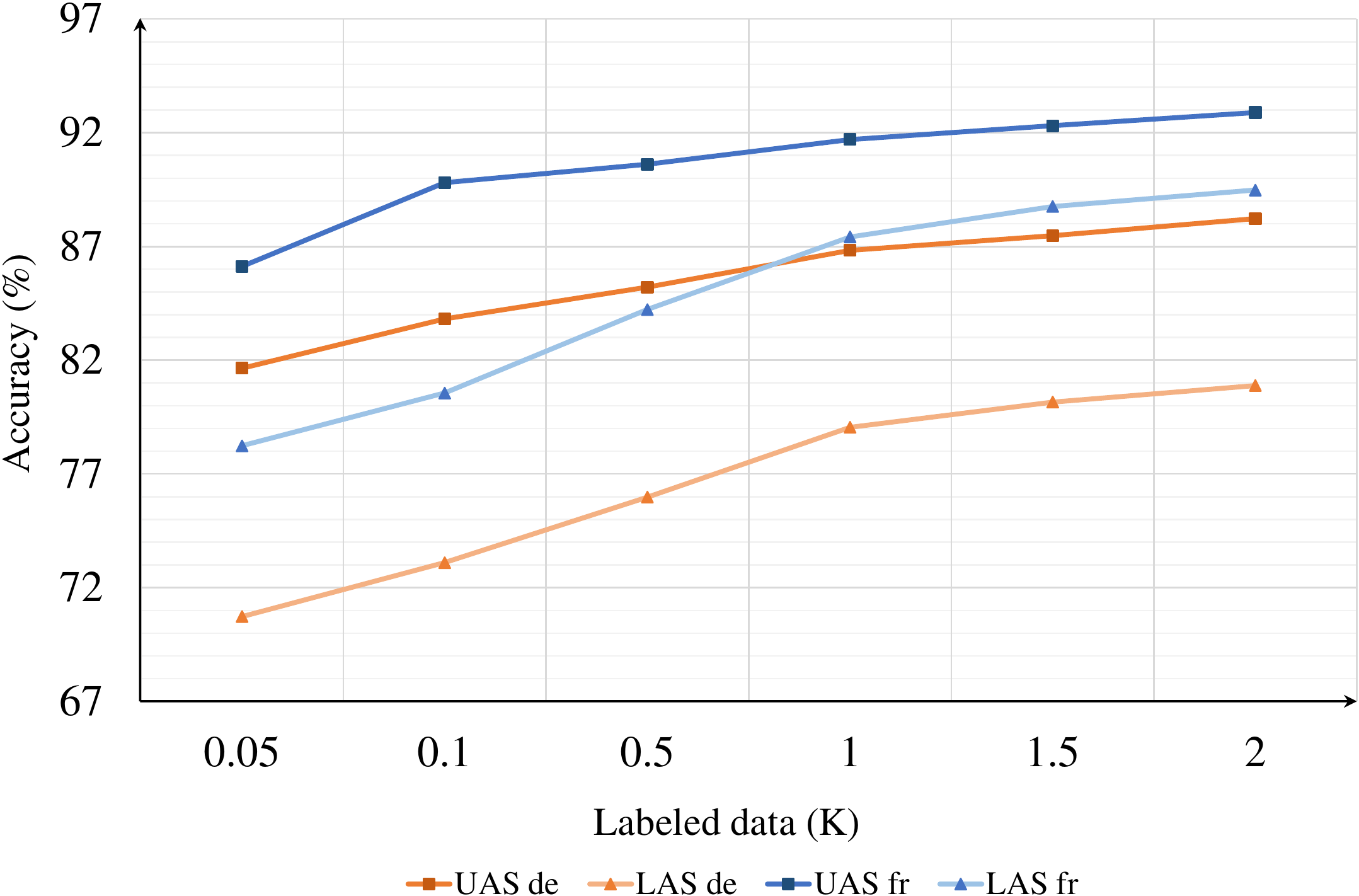}
  \caption{Few-shot cross-lingual learning result on $de$ and $fr$ test sets with different amount of labeled data.}\label{fig:few_shot_self_train_curve}
\end{figure}

We also list the results of our method on the test sets of 22 different UD treebanks in the left hand of Table \ref{tab:ud}, along with the baseline of biaffine dependency parsing according to Ma et al. (2018) \cite{ma-etal-2018-stack}. A common evaluation strategy is adopted for our models that trains the models in five rounds with different random seeds and reports the average results. It demonstrates that our method achieves state-of-the-art results for both UAS and LAS in most languages. In languages with larger training corpus ($cs, ru, es, no$), our method appears to develop more. Moreover, our method tends to improve more in the IE. Romance language family and to improve less or even lower performance in the IE. Germanic language family. This may be caused by the inconsistency of the flexibility of the language order of different language families; WO play a smaller role in the languages with less word order change, so the improvement is relatively less.

With multilingual BERT (mBERT), our model achieves further improvement in most languages, but for some languages ($de, no, nl, ru$), the performance is actually reduced. Since mBERT is a single BERT model trained with Wikipedia to serve 104 languages without any explicit cross-lingual links, it can easily fail to have equally high-quality representation for all languages. This phenomenon has been confirmed by Wu and Dredze (2020) \cite{wu-dredze-2020-languages}.
As the results of $de$ and $nl$ using mBERT decreased too much compared with those without mBERT, we replaced the results of $de$ and $nl$ in the table with the results of using their monolingual BERT. Our average result for the languages selected by Li et al.(2020) \cite{li2020global} is 94.01/91.15 which is a bit lower than 94.55/91.59 in their work. However, since they used UD v2.3, whereas we use 2.2 to be consistent with Ahmad et al. (2019a) \cite{ahmad-etal-2019-difficulties}, the results are incomparable essentially. 

\subsection{Few-shot Effects}\label{sec:few_shot}

We incorporate various quantities of labeled target language data into the training set for few-shot cross-lingual transfer learning, and we use additional unlabeled data of size 10K at the same time. The labeled data comes from the target language training set of UD. The result of this part is shown in Figure \ref{fig:few_shot_self_train_curve}. We can see that, compared to zero-shot, even a small amount of labeled data can bring a significant improvement to the model, especially for LAS, and the accuracy becomes higher and higher with the increase in the amount of labeled data. This is not only because the labeled data directly gives the syntactic information of the target language to the model, but also because the labeled data has a higher degree of similarity to the test set than Wikipedia, so the language modeling of the labeled data is more helpful for cross-lingual transferring. In addition, with the increase of labeled data, the performance growth for few-shot learning gradually decreases, indicating that the improvement brought by the increase of data is not linear. On the one hand, with the introduction of large-scale labeled data, the labeling noise problem has become non-negligible; on the other hand, the capability of the model is also known as an important factor limiting performance.

\begin{table}[]
    \caption{Ablation results on PTB and CTB.}\label{tab:ablation_ptb_ctb}
    \centering
        \begin{tabular}{l|cc|cc}
        \toprule
        \multirow{2}{*}{Systems} & \multicolumn{2}{c|}{\bf PTB} & \multicolumn{2}{c}{\bf CTB} \\ 
        & UAS & LAS & UAS & LAS \\
        \midrule
        Ours  & \textbf{96.05} & \textbf{94.43} & \textbf{91.22} & \textbf{90.05}  \\ 
        Ours w/o WO  & 95.95 & 94.26 & 91.04 & 89.87  \\ 
        Ours w/o MLM  & 95.92 & 94.29 & 90.97 & 89.85 \\ 
        Ours w/o WO \& MLM & 95.83 & 94.19 & 90.32 & 89.07 \\ 
        Ours pipeline & 95.80 & 94.14 & 90.70 & 89.34 \\ 
        \bottomrule
        \end{tabular}
\end{table}

\subsection{Ablation}

Table \ref{tab:ablation_ptb_ctb} shows the monolingual ablation results on PTB and CTB. When only WO or MLM is used as the auxiliary objective, the performance of the model is improved compared with that of the original model, while the model performs best when both two objectives are added at the same time. In addition, to demonstrate the efficacy of our multi-task mode, we use the pipeline mode to train a model instead of multi-tasking. The pipeline mode is to train the model for WO and MLM objectives first and then train a parser based on this model. The result shows that the pipeline mode does little to improve performance and the necessity of multi-task is essential for better incorporating our introduced language modeling objectives. 
In the current boom of pre-training , using MLM to pre-train the encoder and then apply this encoder to downstream tasks is a common enhancement approach. In our proposed method, further explicit combining the objectives of MLM and WO with the downstream task training can also bring further improvement.

\begin{table}[]
    \caption{Zero-shot cross-lingual transferring results.}\label{tab:cross_ablation_de}
    \centering
        \begin{tabular}{l|cc|cc}
        \toprule
        \multirow{2}{*}{Systems} & \multicolumn{2}{c|}{$de^{first}$} & \multicolumn{2}{c}{$de^{final}$} \\ 
        & UAS & LAS & UAS & LAS \\
        \midrule
        Ours  & \textbf{71.16} & 59.28 & \textbf{76.20} & \textbf{63.39} \\ 
        Ours w/o WO  & 70.79 & 59.03 & 75.81 & 63.12 \\ 
        Ours w/o MLM  & 70.82 & 59.16 & 75.76 & 63.24 \\ 
        Ours w/o WO \& MLM & 70.23 & 58.75 & 75.44 & 62.85 \\
        Ours w/o source LM & 71.02 & \textbf{59.31} & 75.74 & 63.15 \\ 
        \midrule
        \multirow{2}{*}{Systems} & \multicolumn{2}{c|}{$fr^{first}$} & \multicolumn{2}{c}{$fr^{final}$} \\ 
        & UAS & LAS & UAS & LAS \\
        \midrule
        Ours  & \textbf{79.01} & \textbf{68.39} & \textbf{83.15} & \textbf{73.23} \\ 
        Ours w/o WO  & 78.71 & 68.18 & 82.77 & 72.37 \\ 
        Ours w/o MLM  & 78.82 & 68.09 & 82.69 & 72.48 \\ 
        Ours w/o WO \& MLM & 78.50 & 67.76 & 82.59 & 72.31 \\ 
        Ours w/o source LM & 78.96 & 68.27 & 82.83 & 72.95 \\
        \bottomrule
        \end{tabular}
\end{table}

Table \ref{tab:cross_ablation_de} shows the zero-shot cross-lingual transfer learning result from $en$ to $de$ and $fr$ in the first round and the final round of self-training in different systems. Our method with both Wo and MLM improves the cross-lingual precision by only using the source language training data, and the final result of our method with both two objectives is also the best. Moreover, the results obtained by excluding source language modeling are lower than the results obtained by including source language modeling, especially after multiple rounds of self-training. We think the reasons are as follows. On the one hand, language modeling on the source language data can help the model better learn the parsing treebank on the source language and gain stronger parsing ability. Moreover, the model is based on the parsing treebank of the source language which differs from the target treebank, so adding the component of the source language into language modeling can improve the transfer learning ability of the model. On the other hand, the amount of data in the source language tends to be smaller than the amount in the unlabeled target language that we use, so modeling the source language data does not interfere much with the modeling of the target language, but instead introduces some perturbations that facilitate cross-lingual transfer learning.

\begin{figure}[]
  \centering
  \includegraphics[scale=0.36]{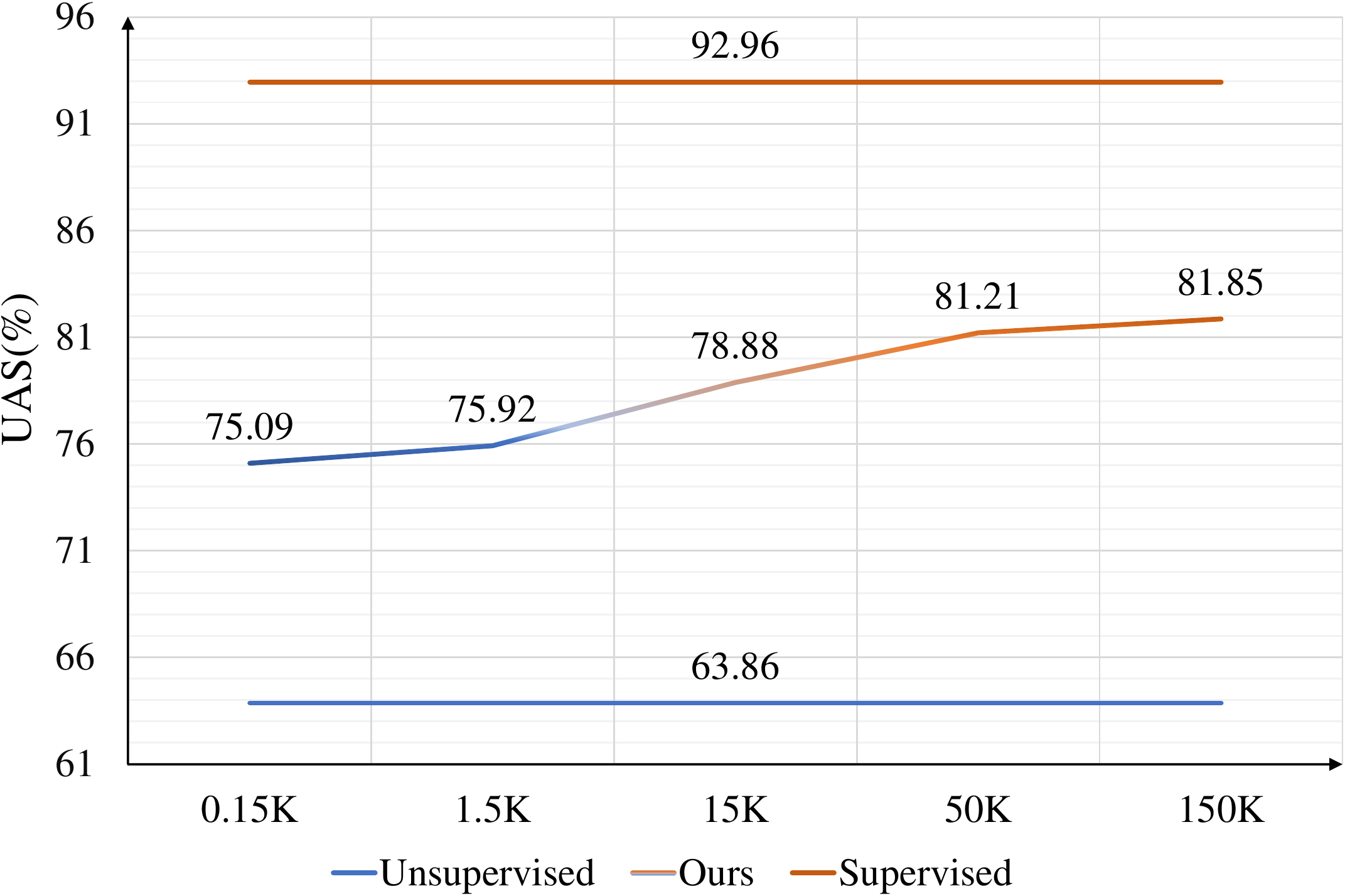}
  \caption{The average growth curve of UAS on UD test set compared with unsupervised and supervised learning.}\label{fig:curve_uas}
\end{figure}

\begin{table}[]
    \caption{Cross-domain transferring results.}\label{tab:cross_domain}
    \centering
    \scalebox{0.9}{
        \begin{tabular}{l|cc|cc|cc|cc}
        \toprule
        \multirow{2}{*}{en} & \multicolumn{2}{c|}{EWT} & \multicolumn{2}{c|}{GUN} & \multicolumn{2}{c|}{LinES} &
        \multicolumn{2}{c}{ParTUT} \\ 
        & UAS & LAS & UAS & LAS & UAS & LAS & UAS & LAS\\
        \midrule
        Ours & 92.06 & 89.95 & 91.18 & 84.20 & 88.64 & 84.28 & 91.04 & 86.17 \\ 
        BIAF & 91.91 & 89.82 & 89.76 & 83.22 & 87.15 & 82.86 & 89.58 & 85.44 \\ 
        \midrule
        \multirow{2}{*}{fr} & \multicolumn{2}{c|}{GSD} & \multicolumn{2}{c|}{Sequoia} & \multicolumn{2}{c|}{Spoken} &
        \multicolumn{2}{c}{ParTUT} \\ 
        & UAS & LAS & UAS & LAS & UAS & LAS & UAS & LAS\\
        \midrule
        Ours & 93.73 & 90.46 & 91.08 & 83.87 & 77.73 & 63.28 & 91.25 & 84.12 \\ 
        BIAF & 92.62 & 89.51 & 90.12 & 83.41 & 72.54 & 60.08 & 90.26 & 85.77 \\ 
        \bottomrule
        \end{tabular}
        }
\end{table}

To examine the potential of using more unlabeled data, we used different amounts of unlabeled data with the size of 0.15K,1.5K,15K,50K and 150K, respectively, to train our model from $en$ to $nl,fr,de,it$ and $es$. Figure \ref{fig:curve_uas} shows the average UAS of this languages. The top red line is the monolingual supervised learning result according to Ma et al. (2018) \cite{ma-etal-2018-stack} and the bottom blue line is the average unsupervised result reported in Yang et al.(2020) \cite{yang-etal-2020-second}. Even using only 0.15K of unlabeled data in the target language, our result is much higher than unsupervised result and with the increase of the amount of unlabeled data, our accuracy keeps approaching the supervised result. There is no strong constraint on the condition that we choose unlabeled data, so the whole 150K data contains many sentences that are not helpful for training. If the unlabeled data are of higher quality, such as containing more complete sentences or containing more sentences similar to those in the target language test set, our results would be better. It is believed that when we obtain enough and high-quality unlabeled data, the performance of our model will be comparable to the results of monolingual supervised learning.

\section{Further Exploration}

\subsection{Cross-domain Effects}
We conducted the experiment on languages en and fr to verify the cross-domain transferring effects of our proposed method. The amount of unlabeled data that we use for each domain is 15K. For $en$, we set $EWT$ as the source treebank and other domain ($GUN,LinES,ParTUT$) as the target treebanks. For $fr$, we set $GSD$ as the source treebank and other domain ($Sequoia,Spoken,ParTUT$) as the target treebank. As shown in Table \ref{tab:cross_domain}, our models perform better than the biaffine dependency parser and has the comparable results to the supervised model trained on its own domain. We do not distinguish between unlabeled data for different domains, so in the case of cross-domains, our model still models the whole language. Therefore, it can be seen that the language modeling objectives in our model can extract universal information for every domain of a certain language. This experiment reveals that our proposed method can not only be used in cross-lingual scenarios due to differences in vocabulary, word order and grammar etc., but also in scenarios where syntax trees are inconsistent due to expression habits in different domains, which further illustrates the adaptability of our proposed method.

\begin{table}[]
    \caption{Cross-lingual results with $fr$ or $ru$ as the source language.}\label{tab:source_effect}
    \centering
        \begin{tabular}{l|cc|cc}
        \toprule
        Lang & \multicolumn{2}{c|}{\bf fr} &  \multicolumn{2}{c}{\bf ru} \\
        & UAS & LAS & UAS & LAS \\
        \midrule
        bg & 83.28 & 71.27 & 89.53 & 78.12 \\
        ca & 88.03 & 78.63 & 74.31 & 60.51 \\
        cs & 74.16 & 59.74 & 77.21 & 65.35 \\
        nl & 72.38 & 65.77 & 73.73 & 63.20 \\
        et & 66.33 & 56.29 & 62.51 & 53.27 \\
        fi & 71.92 & 61.80 & 67.17 & 56.85 \\
        de & 73.97 & 66.29 & 74.03 & 63.95 \\
        it & 91.52 & 86.51 & 77.45 & 68.37 \\
        no & 79.87 & 71.16 & 75.19 & 62.85 \\
        pl & 85.37 & 70.42 & 86.88 & 73.82 \\
        ro & 78.38 & 69.55 & 75.23 & 59.37 \\
        sk & 77.85 & 63.10 & 81.14 & 72.63 \\
        es & 87.82 & 79.08 & 74.95 & 65.72 \\
        \midrule
        AVG & 79.30 & 69.20 & 76.10 & 64.92 \\
        \bottomrule
        \end{tabular}
\end{table}

\subsection{Influence of Loss Weight Bias}
\begin{figure}[]
  \centering
  \includegraphics[scale=0.36]{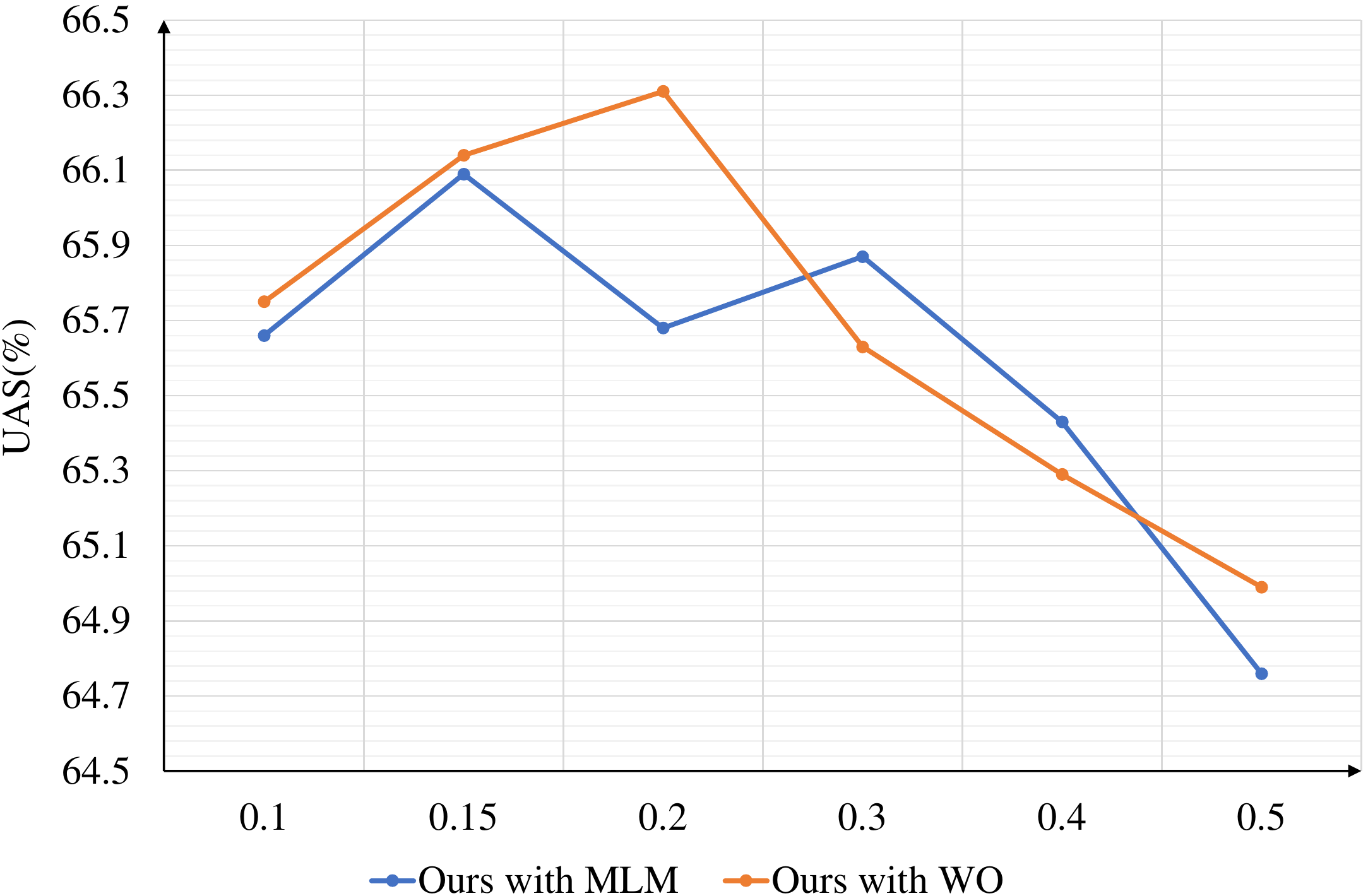}
  \caption{Average UAS after two rounds vs. different loss weight}\label{fig:weight}
\end{figure}

In this section, we perform a hyper-parameter grid search for choosing the suitable loss weight for WO and MLM. We use $en$ as the source language and other 21 languages as the target languages. We use different loss weight from 0.1 to 0.5 to train our model which contains only one of WO and MLM. The average UAS after two rounds of self-training is considered as the criterion for choosing weight. The result is shown in Figure \ref{fig:weight}. Due to the limitation of the size of the training set, our language modeling objectives cannot fully converge. The change curve in the figure shows that, too large loss weight will interfere with the model learning syntactic parsing. While too small loss weight will not make language modeling a valuable contribution to the model. Moreover, under different conditions (w/ MLM, w/ WO), the optimal loss balance weight is different. This is due to the introduction of different objectives, the optimal convergence state changes, so the weights need to be fine-tuned accordingly.

\subsection{Effects of Source Language}
We carry out the experiment to explore the effects of source language. We set fr and ru as the source language respectively and set other languages($bg,ca,cs,de,es,et,fi,it,nl,no,pl,sk,ro$) as the target languages and the results are shown in Table \ref{tab:source_effect}. As we can see in the table, when we used $fr$ as the source language, the model performed better on $bg, ca, es$ and $it$, which cover almost of the languages we chose that were in the same language family as $fr$ (IE.Romance), and when we used $ru$ as the source language, the model performed better on $bg, pl$ and $sk$, most of which belong to the language family of $ru$ (IE.Slavic). We can draw the conclusion that choosing languages of the same family is better for cross-lingual transfer learning. In addition, for languages whose families are not either IE.Romance or IE.Slavic, there is also  a big difference in the performance of $fr$ and $ru$, suggesting the distance between languages are not only determained by families. Therefore, it is rather crude to directly use language family as the criterion to set the value of $\alpha_c$ and $\beta_c$. More precise settings can be inspired by The World Atlas of Language
Structures (WALS) \cite{wals} which provides a great reference for quantifying language distance.

\section{Conclusion}
In this work, we propose a cross-lingual parser that effectively exploits both auxiliary language modeling training objectives and self-training in terms of multi-task mode. Thus we report the first cross-lingual parser only from one source monolingual treebank which is capable of parsing as nearly accurately as the parsers trained on local target treebanks. The results in 21 UD treebanks show that our cross-lingual parser gives an average of 80\% UAS, while the state-of-the-art locally trained parser is 93\% so that this work sparks the first light of accurately parsing all human languages only from one annotated treebank.


\ifCLASSOPTIONcaptionsoff
  \newpage
\fi

% \appendix
%\section{Treebanks Choosing}\label{sec:treebanks}



\bibliographystyle{IEEEtran}
\bibliography{main}

\end{document}